\documentclass[lettersize,journal]{IEEEtran}

\usepackage{amsmath,amsfonts}
\usepackage{algorithmic}
\usepackage{algorithm}
\usepackage{array}
\usepackage[caption=false,font=normalsize,labelfont=sf,textfont=sf]{subfig}
\usepackage{textcomp}
\usepackage{stfloats}
\usepackage{url}
\usepackage{verbatim}
\usepackage{graphicx}
\usepackage{cite}
\usepackage{multirow}
\usepackage{xcolor}
\usepackage{fancyhdr}

\begin{document}

\title{Selective Transfer Learning of Cross-Modality Distillation for Monocular 3D Object Detection}

\author{Rui Ding,~\IEEEmembership{Student Member,~IEEE,} Meng Yang,~\IEEEmembership{Member,~IEEE,} and Nanning Zheng,~\IEEEmembership{Fellow,~IEEE}

\thanks{

Rui Ding, Meng Yang, and Nanning Zheng are with the Institute of
Artificial Intelligence and Robotics, Xi’an Jiaotong University, Xi’an 710049,
P.R.China. E-mails: dingrui@stu.xjtu.edu.cn; \{mengyang, nnzheng\}@mail.xjtu.edu.cn.

Corresponding authors: M. Yang and N. Zheng. 
}

}




\maketitle

\thispagestyle{fancy}

\lfoot{}


\renewcommand{\headrulewidth}{0mm}


\begin{abstract}
Monocular 3D object detection is a promising yet ill-posed task for autonomous vehicles due to the lack of accurate depth information. Cross-modality knowledge distillation could effectively transfer depth information from LiDAR to image-based network. However, modality gap between image and LiDAR seriously limits its accuracy. In this paper, we systematically investigate the negative transfer problem induced by modality gap in cross-modality distillation for the first time, including not only the architecture inconsistency issue but more importantly the feature overfitting issue. We propose a selective learning approach named \textbf{MonoSTL} to overcome these issues, which encourages positive transfer of depth information from LiDAR while alleviates the negative transfer on image-based network. On the one hand, we utilize similar architectures to ensure spatial alignment of features between image-based and LiDAR-based networks. On the other hand, we develop two novel distillation modules, namely Depth-Aware Selective Feature Distillation (\textbf{DASFD}) and Depth-Aware Selective Relation Distillation (\textbf{DASRD}), which selectively learn positive features and relationships of objects by integrating depth uncertainty into feature and relation distillations, respectively. Our approach can be seamlessly integrated into various CNN-based and DETR-based models, where we take three recent models on KITTI and a recent model on NuScenes for validation. Extensive experiments show that our approach considerably improves the accuracy of the base models and thereby achieves the best accuracy compared with all recently released SOTA models.  The code is released on
{\color{blue} https://github.com/DingCodeLab/MonoSTL}.
\end{abstract}


\section{Introduction}

\IEEEPARstart{T}{hree}-dimensional (3D) object detection is a crucial component for scene perception of autonomous vehicles. The dominant solution of 3D detection often relies on LiDAR sensors \cite{shi2019pointrcnn, lang2019pointpillars,yuan2021temporal, zhao2021transformer3d, sugimura2019three}, which has achieved great success in the past decade. However, high cost and complex configuration of LiDAR sensors limit the deployment in practical applications. Monocular 3D detection offers a convenient and flexible solution in recent years \cite{sheng2023pdr,yang2023mix,tao2023pseudo,ma2021delving,liu2022learning,zhang2022monodetr}. However, these methods are limited due to the ill-posed nature of monocular 3D detection. As a result, there is still a huge performance gap between LiDAR-based and monocular 3D detectors due to the lack of accurate depth information. 

\addtocounter{footnote}{-1}

\begin{figure*}[t!]
    \centering
    \includegraphics[width=18cm]{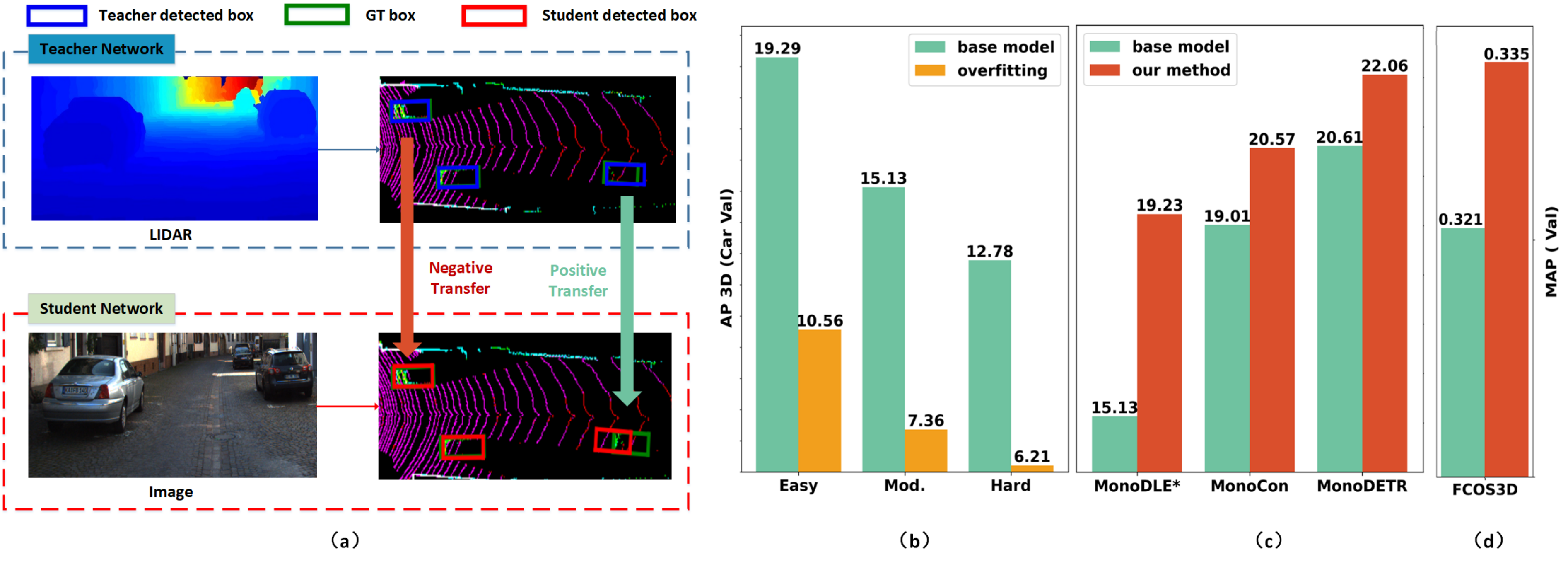}
    \caption{
    \textbf{Negative transfer problem in cross-modality distillation.} (a) A visual example of positive/negative transfer. (b) The accuracy of monocular 3D detection is seriously decreased when fully transferring the features to student from teacher with a similar architecture based on MonoDLE* \protect \footnotemark. Our approach uses selective transfer learning to alleviate the negative transfer problem and considerably improves the accuracy of four base models on (c) KITTI and (d) NuScenes datasets.
    }
    \label{fig:schematic}
\end{figure*}

Recently, cross-modality knowledge distillation provides an innovative solution to effectively leverage LiDAR data for monocular 3D object detection \cite{gao2022esgn}. The teacher network takes LiDAR or a fusion of LiDAR and images as inputs, while the student network uses only images. In the training stage, accurate depth information of LiDAR is transferred to the student network through distillation modules. In the inference stage, only the student network is retained to predict 3D objects from single images. However, it arises two challenging issues due to modality gap between LiDAR and images in cross-modality distillation:

\begin{enumerate}
    \item \textbf{Architecture inconsistency}. The network architectures of LiDAR-based and image-based detectors are generally inconsistent due to different input modalities \cite{cao2022pkd}. LiDAR-based detectors generally use point-based \cite{qi2017pointnet} or voxel-based \cite{zhou2018voxelnet} networks, while image-based detectors use CNN-based \cite{zhou2019objects} or transformer \cite{zhang2022monodetr} architectures. Their architecture inconsistency incurs extra burden to handle spatially-unaligned intermediate features between the teacher and student networks. 
    \item \textbf{Feature overfitting}. Learned features of LiDAR-based and image-based detectors also differ due to different input modalities. LiDAR contains precise depth information, while images contain texture and color details \cite{liu2023bevfusion,li2022deepfusion}. The student network tends to overfit to features of the teacher network during training, because accurate depth information is beneficial to high prediction accuracy. However, learned features of the student network may not be effective enough due to the lack of accurate depth information during inference. 
\end{enumerate}

These two issues are collectively referred as the \textit{negative transfer} problem of cross-modality distillation in our work. That means, learned features from the teacher network are not always beneficial to the student network.
The fundamental reason behind the negative transfer problem is the modality gap between LiDAR and images.
PKD \cite{cao2022pkd} already confirmed that the architecture inconsistency issue may lead to performance decline for the student network.
Fig. \ref{fig:schematic}(a) shows visual examples of positive and negative transfer induced by the feature overfitting issue. When the student network already localizes 3D objects accurately compared to ground-truth (GT), transferring the features from teacher to student may worsen the localization accuracy, and vice versa.
Fig. \ref{fig:schematic}(b)-(d) quantitatively verifies that the negative transfer problem has a significant impact in cross-modality distillation and our selective transfer learning approach well alleviates this problem on both the KITTI and NuScenes datasets.

A few methods in recent years made preliminary attempts on the negative transfer problem. On the one hand, the architecture inconsistency issue is relatively easy to avoid. For instance, Monodistill \cite{chong2022monodistill} and ADD \cite{wu2023attention} used similar network architectures for the teacher and student networks, while CMKD \cite{hong2022cross} conducted distillation from a Bird's Eye View (BEV) perspective to alleviate the architecture inconsistency issue partly. On the other hand, the feature overfitting issue is more challenging and not well investigated in the literature. The methods \cite{chong2022monodistill,wu2023attention} simply distinguished foreground and background to filter away noises in the background, which partly alleviated the feature overfitting issue. 
Similar problems were studied in the field of transfer learning {\cite{zhang2022survey}}. However, traditional transfer learning focuses on transferring knowledge between different tasks or domains {\cite{pan2009survey,tan2018survey}}, while our task deals with transferring knowledge between different modalities.
In conclusion, existing methods did not fully handle the negative transfer problem of cross-modality distillation in monocular 3D detection, especially for the feature overfitting issue.

\footnotetext{MonoDLE* refers to the base model in Monodistill \cite{chong2022monodistill}, which is developed on the basis of MonoDLE \cite{ma2021delving}. }

\addtocounter{footnote}{1}

\footnotetext{FCOS3D* refers to its open-source model, which is developed on the basis of FCOS3D \cite{wang2021fcos3d}. }

\addtocounter{footnote}{-1}

In this paper, we systematically investigate the negative transfer problem caused by modality gap in cross-modality distillation including not only the architecture inconsistency issue but more importantly the feature overfitting issue. We propose a selective learning approach named \textbf{MonoSTL} to well alleviate the negative transfer problem in monocular 3D object detection. Our approach utilizes similar network architectures for the teacher and student networks following \cite{cao2022pkd} to address the architecture inconsistency issue, while develops two novel distillation modules to address the feature overfitting issue. In this way, our approach encourages positive transfer of depth information from LiDAR while alleviates the negative transfer on image-based network.

First, we introduce depth uncertainty as a criterion to selectively learn depth information from the teacher network. In cross-modality distillation, LiDAR-based network supplements accurate depth information that is missed in images. Depth uncertainty could well reflect the depth prediction capability of the student network. Therefore, we use it to determine the proportion of depth information from the teacher network in the training. Second, we develop two novel distillation modules, namely, Depth-Aware Selective Feature Distillation (\textbf{DASFD}) and Depth-Aware Selective Relation Distillation (\textbf{DASRD}). These two modules distinguish the importance of objects during the distillation process and emphasize the selective learning of features in the teacher network. Specifically, \textbf{DASFD} and \textbf{DASRD} selectively learn positive features/relationships of objects from the teacher network by integrating depth uncertainty into feature/relation distillation.

Our approach can be seamlessly integrated into various CNN-based and DETR-based models without increasing inference costs. In this paper, it is comprehensively applied on four recent open-source models for easy implementation including MonoDLE*, MonoCon \cite{liu2022learning}, MonoDETR, and FCOS3D* \protect \footnotemark. The results demonstrate that our approach considerably improves the accuracy of the base models on KITTI in Fig. \ref{fig:schematic}(c) and NuScenes in Fig. \ref{fig:schematic}(d). It thereby achieves the best accuracy compared with all released SOTA models in recent years.
In addition, our approach holds the potential to achieve higher accuracy with stronger models in the future.

Our contributions are highlighted as follows:

\begin{itemize}
    \item We systematically investigate the negative transfer problem of cross-modality distillation in monocular 3D detection for the first time.
    \item We propose two novel distillation modules to alleviate the problem, which can be seamlessly integrated into various CNN-based and DETR-based models.
    \item Our approach considerably improves the accuracy of four recent models on two datasets. It thereby achieves the best accuracy compared with all released SOTA models.

\end{itemize}

\begin{figure*}[t!]
    \centering
    \includegraphics[width=17cm]{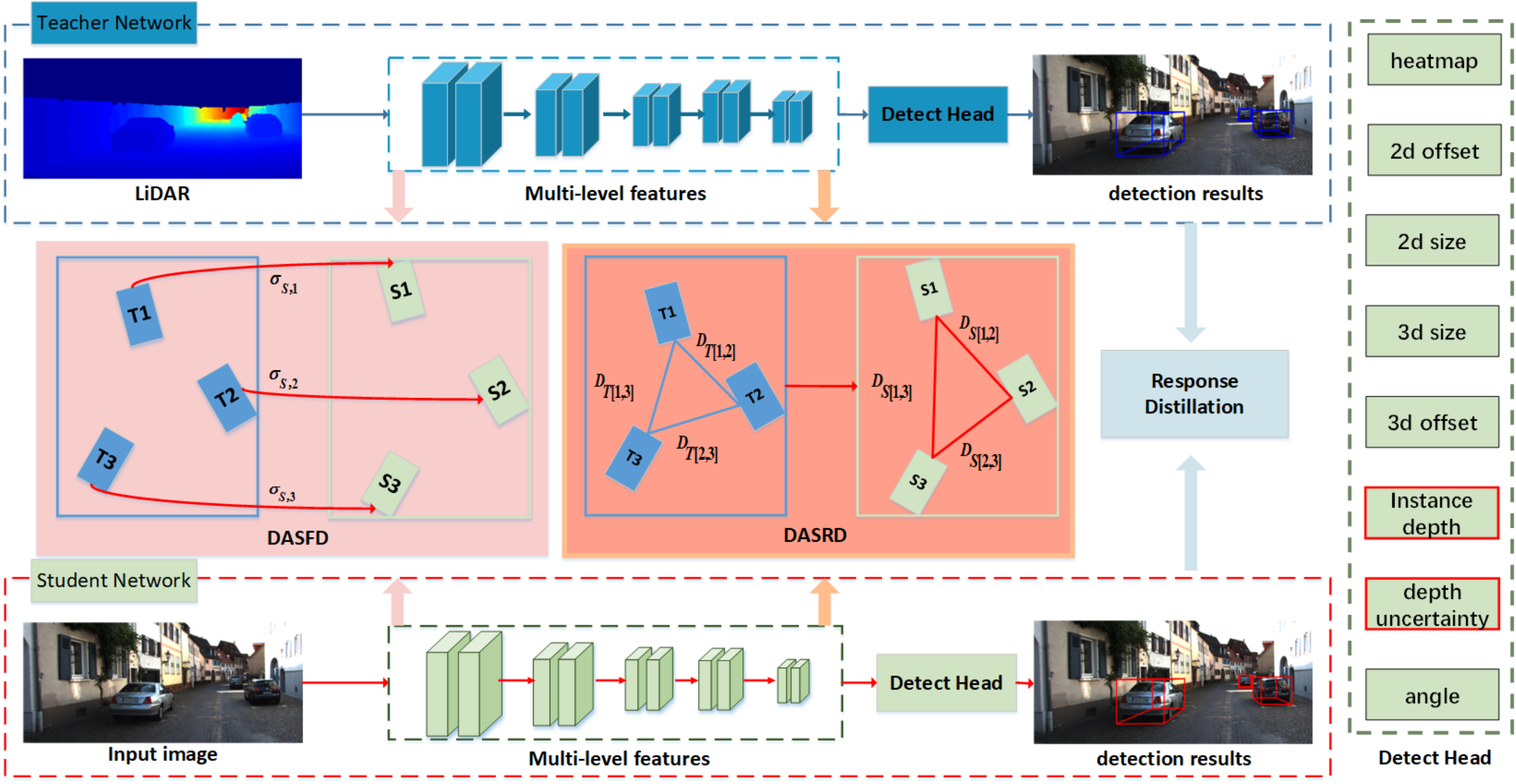}
    \caption{\textbf{Overview of our MonoSTL framework}. The framework comprises three components including the teacher network, the student network, and three distillation modules. First, we train the teacher network with GT depth map from LiDAR. It adopts similar architectures to the student network. Second, three distillation modules are used to selectively transfer features from the teacher network to the student network including our DASRD and DASFD modules as well as the general response distillation module. Finally, only the student network is retained to predict 3D objects from single images in the inference stage. 
}
    \label{fig:framework}
\end{figure*}

\section{Related work}

\subsection{Monocular 3D Object Detection}
Monocular 3D detection predicts locations, dimensions, and angles of 3D objects from single images. Due to the lack of accurate depth information in images, there is still a huge performance gap between monocular 3D detection and LiDAR-based 3D detection. MonoDLE \cite{ma2021delving} verified that the accuracy of monocular 3D detection is mainly limited by depth prediction compared to LiDAR-based 3D detection. Recent approaches can be divided into two categories based on whether extra depth data is involved in the training. Monocular 3D detection without extra depth data often predicts depth information from single images through 2D-3D geometric constraints and prior knowledge. For instance, GupNet \cite{lu2021geometry} used height priors of objects for depth estimation. MonoFlex \cite{zhang2021objects} combined multiple depth estimation methods using depth uncertainty. MonoDDE \cite{li2022diversity} introduced a keypoint-based method to further improve the accuracy of depth estimation. With the success of Detection Transformer (DETR) \cite{carion2020end} in 2D detection, MonoDETR \cite{zhang2022monodetr} and MonoDTR \cite{huang2022monodtr} extended DETR to monocular 3D detection. However, these methods are limited due to the ill-posed nature of monocular 3D detection. The methods with extra depth data have attracted considerable attention in recent years. Pseudo-LiDAR \cite{wang2019pseudo,weng2019monocular,wang2020task} converted depth maps into pseudo-LiDAR and utilized LiDAR detectors for 3D object detection. However, these methods do not well utilize LiDAR data. In this paper, we adopt cross-modality distillation to effectively leverage depth information from LiDAR data in the training phase for monocular 3D detection.

\subsection{Cross-Modality Knowledge Distillation}
Knowledge distillation was originally proposed for model compression \cite{hinton2015distilling} and then widely applied in various domains \cite{liu2021deep,ma2023using}. Most knowledge distillation approaches use the same input modality for the teacher and student networks, while cross-modality knowledge distillation involves different input modalities. 2DPASS \cite{yan20222dpass} used images to boost the representation learning on LiDAR. Monodistill \cite{chong2022monodistill} and ADD \cite{wu2023attention} utilized LiDAR as inputs for the teacher network to enhance the depth prediction capability of the student network. Monodistill and ADD used similar network architectures for the
teacher and student network, while CMKD \cite{hong2022cross}, BEVDistill \cite{chen2022bevdistill} and UniDistill \cite{zhou2023unidistill} conducted distillation from BEV perspective to partly alleviate the architecture inconsistency issue in cross-modality distillation. However, these methods do not fully consider the negative transfer problem induced by modality gap in cross-modality distillation, particularly for the feature overfitting issue. In this paper, we systematically investigate and alleviate the negative transfer problem for the first time, including not only the architecture inconsistency issue but more importantly the feature overfitting issue.

\subsection{Depth Uncertainty}
Depth uncertainty refers to the confidence in the accuracy of perceived depth information. It is crucial in autonomous driving to ensure safe and reliable decision-making, especially in complex and dynamic environments. Therefore, depth uncertainty estimation is as important as depth estimation for 3D object detection. A fundamental formula of depth uncertainty estimation was proposed in \cite{kendall2017uncertainties} and it has been widely adopted in 3D detection such as MonoPair \cite{chen2020monopair} and GupNet. For example, MonoFlex and MonoDDE leveraged depth uncertainty to combine various depth estimation methods and employed a soft fusion approach to enhance depth values. To our knowledge, we first utilize depth uncertainty to selectively learn the depth information of the teacher network. By integrating depth uncertainty, our approach well alleviates the negative transfer problem in cross-modality distillation.

\section{Method}

\subsection{Overall Framework}

Fig. \ref{fig:framework} presents the overall framework of our approach. It comprises three components: 1) the student network takes a single image as input, 2) the teacher network adopts a similar architecture but different inputs (either LiDAR or a fusion of LiDAR and image), and 3) two novel distillation modules including Depth-Aware Selective Feature Distillation (\textbf{DASFD}) and Depth-Aware Selective Relation Distillation (\textbf{DASRD}).

The student network could adopt various CNN-based and DETR-based detection models without any modifications. It ensures the seamless compatibility of our approach with various models without increasing inference cost. In this paper, we take three recent models on KITTI to validate our approach including MonoDLE \cite{ma2021delving}, MonoCon \cite{liu2022learning} and MonoDETR \cite{zhang2022monodetr}. In addition, we take a recent model FCOS3D* \cite{wang2021fcos3d} on NuScenes to validate the effectiveness of our approach across various datasets. The teacher network utilizes a similar architecture to alleviate the architecture inconsistency issue. The two novel distillation modules \textbf{DASFD} and \textbf{DASRD} are used to alleviate the feature overfitting issue. \textbf{DASFD} selectively learns positive features of objects from the teacher network by integrating depth uncertainty into feature distillation, while \textbf{DASRD} selectively learns positive relationships of objects by integrating depth uncertainty into relation distillation. In addition, the classical response distillation \cite{chong2022monodistill} is incorporated but not the focus of this work.

In our framework, we define levels of intermediate features obtained from backbones as $\{S^{(l)}\}_{l\in(1,...,L)}$ in the student network and $\{T^{(l)}\}_{l\in(1,...,L)}$ in the teacher network. The total number of intermediate feature levels used for distillation is denoted as $L$. For each object, we obtain the length and width of its feature map using GT 2D bounding box. These bounding boxes well distinguish foreground and background regions and they are used in the ROIAlign operation. Notably, these values are only utilized during the training phase and do not affect the inference process. Specifically, we assign the index $i$ to represent the $i\text{-th}$ objects, and the total number of objects in GT is denoted as $N$. Notably, the intermediate layers utilized for distillation are determined according to the used base models in the student network. 

\subsection{ Depth-Aware Selective Feature Distillation}

In general knowledge distillation \cite{romero2014fitnets,kim2018paraphrasing}, the student network was directly forced to mimic the feature map from the teacher network. It may result in the feature overfitting issue. Therefore, we first introduce the Depth-Aware Selective Feature Distillation (\textbf{DASFD}) to alleviate the feature overfitting issue in cross-modality distillation.

\textbf{1) General Feature Distillation.}
The general feature distillation loss $L_{fd}$ is as follows:

\begin{equation}
L_{fd}=\sum_{{l=1}}^L \frac{1}{H^{(l)} \times W^{(l)}} F\left(T^{(l)}, S^{(l)}\right) ,
\label{general feature distillnation}
\end{equation}
where $H^{(l)}$, $W^{(l)}$ denote the height and width of the feature map for distillation, and $F(\cdot)$ represents different feature distillation functions. In the general knowledge distillation, the student network fully relies on features from the teacher network, leading to effective inference when they share the same input modality and have similar network architectures. However, in cross-modality knowledge distillation, the modality gap between LiDAR and images may lead to feature overfitting of the student network to the teacher network. The feature overfitting issue results in ineffective features of the student network during inference. To address this issue, 
we develop a selective learning approach, which distills only positive information while avoids interference from negative information in cross-modality distillation.

\textbf{2) Depth Uncertainty.}
First of all, we require an effective metric to achieve selective learning.
Fig. 2 shows that the detect head modules in both the teacher and student networks predict instance depth, depth uncertainty, and other prediction heads in cross-modality distillation.
LiDAR-based network supplements accurate depth information that is missed in image-based network.
Therefore, depth uncertainty could well reflect the depth prediction capability of the student network. It provides a reasonable criterion for our solution. We use it to determine the proportion of knowledge to be learned from the teacher network in the training.
Notably, though the teacher network directly provides pixel-level depth values from LiDAR in the input, both the teacher and student networks still require to predict instance depth and depth uncertainty for objects.

Similar to previous methods, we adopt the classical depth estimation formula in \cite{kendall2017uncertainties}, which obtains instance depth and depth uncertainty simultaneously. The expression of the depth loss $L_{dep}$ is as follows:

\begin{equation}
L_{\text {dep }}=\frac{\left|z-z^*\right|}{\sigma}+\log (\sigma) ,
\label{depth uncertainty loss}
\end{equation}
where $z$ and $z^\ast$ represent predicted depth value and ground truth (GT), respectively. The symbol $\sigma$ denotes the depth uncertainty. Both $z$ and $\sigma$ can be predicted from our networks. The second term in the formula prevents infinite output values of networks, yielding reasonable depth uncertainty to reflect the quality of depth prediction \cite{su2022uncertainty}. Higher depth uncertainty indicates that predicted depth values of the student network are more unreliable. Therefore, more depth information of the teacher network should be transferred to the student network, and vice versa.

In our solution, depth uncertainty is obtained from the student network. Specifically, we calculate a weight $\omega_i$ as $ \omega_i= \sigma_{S, i}$ for the $i\text{-th}$ object in the student network.
In this way, the weight value directly indicates the accuracy of predicted objects. When the student network accurately predicts objects, the distillation weight value is reduced to alleviate the interference of the features from the teacher network. On the contrary, the distillation weight value is increased to transfer more depth information to the student network. Notably, depth uncertainty can be optionally obtained from the teacher network or a fusion of the teacher and student networks. The ablation study of different weight schemes can be seen in Table \ref{Comparisons different weight scheme}.

\textbf{3) Our Feature Distillation.}
Our \textbf{DASFD} module utilizes the weighted depth uncertainty of objects to optimize the cross-modality feature distillation loss, denoted as $L_{wfd}$, as follows:

{\small
\begin{equation}
L_{wfd}=\sum_{l=1}^L \sum_{i=1}^N M_i^{(l)} \omega_i^{(l)} \frac{1}{H_i^{(l)} \times W_i^{(l)}} F\left(T_i^{(l)}, S_i^{(l)}\right) ,
\label{DASFD}
\end{equation}
}where $ H_i^{(l)}$ and $W_i^{(l)}$ denote the height and width of the $i\text{-th}$ foreground object in feature map, respectively. $T_i^{(l)}$ and $S_i^{(l)}$ represent the feature maps of the $i\text{-th}$ foreground object obtained from the teacher and student networks, respectively. We adopt the simple L2 loss as the fundamental distillation function $F(\cdot)$, and it is calculated pixel-by-pixel between feature maps of the teacher and student networks. Our feature distillation loss differs from the general distillation loss in two aspects. On the one hand, $M_i^{(l)}$ represents the foreground mask of the $i\text{-th}$ object, which is obtained from 2D bounding box ground truth. This approach distinguishes foreground and background to effectively filter away noises in the background. On the other hand, $\omega_i^{(l)}$ represents our selective learning weights for objects in cross-modality distillation. We use an effective metric to distinguish the importance of objects, which allows our approach to selectively learn positive features of objects from the teacher network.

\subsection{Depth-Aware Selective Relation Distillation}

Relative relations of objects can also transfer depth information from the teacher network to the student network \cite{tung2019similarity,peng2019correlation,park2019relational}, which exhibits a low correlation with input modalities. Therefore, we then introduce the Depth-Aware Selective Relation Distillation (\textbf{DASRD}) to further alleviate the feature overfitting issue in cross-modality distillation.

\textbf{1) General Relation Distillation.}
The general relation distillation loss is as follows:

{\small
\begin{equation}
L_{r d}=\sum_{l=1}^L \sum_i^N \sum_j^N G\left(R\left(T_i^{(l)}, T_j^{(l)}\right), R\left(S_i^{(l)}, S_j^{(l)}\right)\right),
\label{general relation distillnation}
\end{equation}
}where $T_i^{\left(l\right)}$ and $S_i^{\left(l\right)}$ represent feature maps of the $i\text{-th}$ foreground object from the teacher and student networks, respectively. Notably, the size of feature maps is fixed by applying the RoIAlign operation \cite{he2017mask}. The function $R(\cdot)$ is employed to calculate the feature similarity between the $i\text{-th}$ and $j\text{-th}$ objects in the teacher and student networks separately. The function $G(\cdot)$ is then utilized to compute the pairwise relationship similarity between the teacher and student networks. Relation distillation transfers structural knowledge by capturing the relationships among objects. The structural knowledge exhibits a low correlation with the input modalities of the networks and it is not susceptible to modality gap. Therefore, it is particularly suitable for cross-modality distillation in 3D object detection.

\textbf{2) Our Relation Distillation.}
Existing relation distillation methods often considered foreground objects equally. However, in cross-modality distillation, the importance of objects may vary significantly. For example, objects that are accurately predicted usually transfer more useful information than others. Based on this, we can categorize objects into positive objects and negative objects. Therefore, the relationships among positive objects are more crucial than those among negative objects. The relationships of positive objects can lead to positive transfer effect in cross-modality distillation. Conversely, the relationships of negative objects may result in severe negative transfer problem.

Motivated by this observation, we propose the Depth-Aware Selective Relation Distillation (\textbf{DASRD}) module to further consider the importance of objects in relation distillation. Similar to the \textbf{DASFD} module, we also utilize depth uncertainty to measure the importance of objects in the relation distillation loss. Specifically, we incorporate depth uncertainty in equation (\ref{depth uncertainty loss}) to the function $R(\cdot)$ to compute the correlation between two objects as follows:

{\small
\begin{equation}
D_{T\left [ i,j \right ] }=\sum_{l=1}^L\left(\frac{R\left(T_i^{(l)}, T_j^{(l)}\right)}{\sigma_{T, i}^{2(l)}+\sigma_{T, j}^{2(l)}}+\log \left(\sigma_{T, i}^{2(l)}+\sigma_{T, j}^{2(l)}\right)\right) ,
\label{weight teacher relation}
\end{equation}
}

{\small
\begin{equation}
D_{S\left [ i,j \right ] }=\sum_{l=1}^L\left(\frac{R\left(S_i^{(l)}, S_j^{(l)}\right)}{\sigma_{S, i}^{2(l)}+\sigma_{S, j}^{2(l)}}+\log \left(\sigma_{S, i}^{2(l)}+\sigma_{S, j}^{2(l)}\right)\right) ,
\label{weight student relation}
\end{equation}
}where $D_{T\left [ i,j \right ] }$ and $D_{S\left [ i,j \right ] }$ denote the correlations between paired objects in the teacher network and the student network, respectively. $\sigma_{T,i}$ and $\sigma_{S,i}$ represent depth uncertainty of the $i\text{-th}$ object estimated by the teacher network and the student network, respectively. The two functions $D_{T\left [ i,j \right ] }$ and $D_{S\left [ i,j \right ] }$ in equation (\ref{weight teacher relation}) and (\ref{weight student relation}) are then used to replace the function $R(\cdot)$ in the relation distillation loss in equation (\ref{general relation distillnation}). Our relation distillation loss, denoted as $L_{wrd}$, can be represented as follows:

\begin{equation}
L_{wrd}=\sum_i^N \sum_j^N G\left( D_{T\left [ i,j \right ] }, D_{S\left [ i,j \right ] } \right) .
\label{our relation distillation loss}
\end{equation}
In our solution, we choose L1 loss as the distillation function $G(\cdot)$. In equation (\ref{weight teacher relation}) and (\ref{weight student relation}), a smaller value of denominator implies a higher importance of their relative relationship. In this case, these objects are assigned higher weights in our loss, and the network pays more attention to the relationships among positive targets. In this way, the student network can selectively learn relative relationships of objects to alleviate the feature overfitting issue in cross-modality distillation.

\subsection{Loss Function}

The total loss function of our model is as follows:
\begin{equation}
L=L_{src}+\lambda_1 \cdot L_{wfd}+\lambda_2 \cdot L_{wrd}+\lambda_3 \cdot L_{ed},
\label{total loss}
\end{equation}
where $L_{wfd}$ represents the feature distillation loss in our \textbf{DASFD} and $L_{wrd}$ represents the relation distillation loss in our \textbf{DASRD}. $L_{src}$ represents the general loss to train 3D detection, which is determined by employed base models such as MonoDLE*, MonoCon, MonoDETR and FCOS3D* in our work.
$L_{ed}$ represents the classical response distillation loss as follows: 

\begin{equation}
L_{ed} = \frac{1}{K} \sum_{\kappa=1}^{K} RE(Y_{T,\kappa},Y_{S,\kappa}) ,
\label{general response distillnation}
\end{equation}
where $K$ represents the number of detection heads. $Y_{T,\kappa}$ and $Y_{S,\kappa}$ represent the features of the $\kappa\text{-th}$ detection head from the teacher and student networks. The detection heads include instance depth, depth uncertainty, and other prediction heads shown in Fig. 2. The function $RE(\cdot)$ is utilized to compute the pairwise features of all these detection heads between the teacher and student networks. We choose L1 loss as $RE(\cdot)$ and calculate these head features pixel-by-pixel. The hyperparameters $\lambda_1$, $\lambda_2$, and $\lambda_3$ are utilized to balance the four losses. The ablation study of different loss weights can be seen in Table \ref{tab:hyperparameters}.

\begin{algorithm}[t!]
	\renewcommand{\algorithmicrequire}{\textbf{Input:}}
	\renewcommand{\algorithmicensure}{\textbf{Output:}}
	\caption{Our feature distillation loss in \textbf{DASFD}}
	\label{alg1}
	\begin{algorithmic}[1]
  	\REQUIRE  GT 2D box $M$, teacher network features           $T$, student network features $S$
 	\ENSURE  our feature distillation loss $L_{fd} $
		\STATE Initialization: $l \leftarrow 0$, student depth uncertainty $\sigma_S \leftarrow 0$, $L \leftarrow (\text{total numbers of feature levels of } S )$, object weight $\omega \leftarrow 0$, $L_{wfd} \leftarrow 0$
		\REPEAT
		\STATE Update $\sigma^{(l)}_S$ based on Equation (\ref{depth uncertainty loss})
		\STATE Update $\omega^{(l)}$  based on $\sigma^{(l)}_S$
		\STATE Update $L_{wfd}$ based on Equation~(\ref{DASFD})
        \STATE $l \leftarrow l+1$
		\UNTIL $l >= L$

	\end{algorithmic}  
\end{algorithm}

\subsection{Pseudocodes of our distillation modules}
The main technical contributions of our approach lie in the two novel distillation modules including \textbf{DASFD} and \textbf{DASRD}. For clarity and easy reproduction, we further present the pseudocodes of the losses in these two modules including our feature distillation loss in Algorithm \ref{alg1} and our relation distillation loss in Algorithm \ref{alg2}. Notably, these two losses are not limited by network architectures, which can be easily integrated into to various base models for cross-modality knowledge distillation in the future.

\section{Experiments}

\subsection{Setting}

\textbf{1) Dataset and Evaluation Metrics.} \ 
Our approach is mainly evaluated on the widely used KITTI dataset\cite{geiger2012we} for a fair comparison. This dataset is the most common benchmark for monocular 3D detection. The dataset consists of 7481 images for training and 7518 images for testing. We divide the training data into a training set with 3712 images and a validation set with 3769 images following \cite{chen20153d}. Similar to previous methods, our approach primarily focuses on detecting objects in the Car category while secondarily detecting Pedestrian and Cyclist. Additionally, objects within the KITTI dataset are categorized into three difficulty levels (i.e. Easy, Moderate, and Hard) based on factors such as occlusion, truncation, and size. The primary metrics used for performance comparison are BEV Average Precision ($AP_{BEV}$) and 3D Average Precision ($AP_{3D}$). Following \cite{simonelli2019disentangling}, 40 recall positions are sampled to calculate AP. Notably, the IOU thresholds are 0.7 for Car, and 0.5 for Pedestrian and Cyclist.
In addition, we evaluate the effectiveness of our approach across datasets on NuScenes. NuScenes \cite{caesar2020nuscenes} is a large-scale multi-modal dataset, including multi-view RGB images from 6 cameras, points from 5 Radars and 1 LiDAR. There are overall 1.4M annotated 3D bounding boxes across 10 categories. We extract single-view data from NuScenes for monocular 3D detection. Evaluation metrics include mean Average Precision (mAP) and nuScenes detection score (NDS).

\textbf{2) Basic Model Setting.} \
Our model is trained on a single NVIDIA 3090 GPU. We adopt similar settings to the base models (MonoDLE*, MonoCon, MonoDETR, and FCOS3D*) including the optimizer, epochs, batch size, and data augmentation. The loss weights $\lambda_1$ and $\lambda_2$ are set as 10 and 1, respectively. The loss weight $\lambda_3$ is set to 1 for CNN-based models and 0.1 for DETR-based models.
We train our model in an end-to-end way similar to \cite{chong2022monodistill}. The student network is independently trained or directly loading pertained weights of base models. The teacher network is independently trained and its parameters are frozen. During distillation, the features of the student network pass through two convolution layers and the loss is computed together with the features of the teacher network based on \textbf{DASFD} and \textbf{DASRD}. In the inference, the student network only employs images as input.

\begin{algorithm}[t!]
	\renewcommand{\algorithmicrequire}{\textbf{Input:}}
	\renewcommand{\algorithmicensure}{\textbf{Output:}}
	\caption{Our relation distillation loss in \textbf{DASRD}}
	\label{alg2}
	\begin{algorithmic}[1]
  	\REQUIRE  GT 2D box $M$, teacher network features           $T$, student network features $S$
 	\ENSURE  our relation distillation loss $L_{rd} $
		\STATE Initialization: $l \leftarrow 0$, depth uncertainty $\sigma_T,\sigma_S \leftarrow 0$, $L \leftarrow (\text{total numbers of feature levels of } S )$, $L_{wrd} \leftarrow 0$, relations between paired objects $D_T,D_S \leftarrow 0$
		\REPEAT
		\STATE Update $\sigma^{(l)}_T,\sigma^{(l)}_S$ based on Equation (\ref{depth uncertainty loss})
		\STATE Update $T,S$ by ROIAlign operation
        \STATE Update $D_T,D_S$ based on Equation~(\ref{weight teacher relation}) and (\ref{weight student relation})
		\STATE Update $L_{wrd}$ based on Equation~(\ref{our relation distillation loss})
        \STATE $l \leftarrow l+1$
		\UNTIL $l >= L$

	\end{algorithmic}   
\end{algorithm}

\subsection{Implementation details on base models}

Our approach is comprehensively verified by integrating into different kinds of base models. Four representative open-source models are adopted for easy implementation including MonoDLE*, MonoCon, MonoDETR and FCOS3D*. MonoDLE* has been widely adopted for monocular 3D detection in recent years, which ensures a fair comparison with other models. MonoCon and MonoDETR are two recent CNN-based and DETR-based models, which achieved SOTA on KITTI among all released models in recent years. FCOS3D* is implemented on the NuScenes dataset.

\textbf{1) Network Structure. }
Both MonoDLE* and MonoCon employ the DLA-34 architecture as the backbone and utilize multiple CNN-based heads to predict 3D object attributes from single images. FCOS3D* employs ResNet-101 as its backbone network and utilizes multiple CNN-based heads. MonoDETR employs ResNet-50 as its backbone network and integrates a visual encoder and a depth encoder to extract image and depth features, respectively. These features are then combined in a depth-aware transformer module to predict 3D detection attributes.

\textbf{2) Inputs and Distillation Layers. }For MonoDLE*, MonoCon and FCOS3D*, the teacher network employs depth maps as input, which are transformed from LiDAR data. For MonoDETR, the teacher network employs a visual encoder for image inputs and a depth encoder for depth maps inputs. It ensures that the depth encoder branch of the teacher network is mainly influenced by input LiDAR data. The layers used for knowledge distillation vary for different network architectures. For MonoDLE*, MonoCon and FCOS3D*, knowledge distillation is performed from the last three layers of the backbone network. For MonoDETR, knowledge distillation is only performed from two feature layers of the depth encoder.

\begin{table*}[t!]
\centering
\caption{\textbf{Comparison to the three base models on KITTI}. Our approach consistently outperforms the three base models across all difficulty levels in both the training and validation sets.}
\setlength{\tabcolsep}{4.0mm}
{
\begin{tabular}{c|ccc|ccc|ccc}
\hline 
\multirow{2}{*}{ Method } & \multicolumn{3}{c|}{$AP_{3D}$ (Car test) } & \multicolumn{3}{c|}{$AP_{3D}$ (Car val) } & \multicolumn{3}{c}{$AP_{BEV}$ (Car val) } \\
\cline{2-10}
&  Easy & Mod. &  Hard  &  Easy  &  Mod.  &  Hard  &  Easy  &  Mod.  &  Hard  \\
\hline 
MonoDLE * \cite{ma2021delving}                    & 17.23 & 12.26 & 10.29 & 19.29 & 15.13 & 12.78 & 26.47 & 20.24 & 18.29 \\
MonoDLE* + \textbf{Ours}      & 24.04 & 16.50 & 13.85 & 25.59 & 19.23 & 16.60 & 34.20 & 25.82 & 22.47 \\
Improvement                  & \textbf{+6.81} & \textbf{+4.24} & \textbf{+3.56} & \textbf{+6.3} & \textbf{+4.1} & \textbf{+3.82} & \textbf{+7.73} & \textbf{+5.58} & \textbf{+4.18} \\
\hline 
MonoCon \cite{liu2022learning}                     & 22.50 & 16.46 & 13.95 & 26.33 & 19.01 & 15.98 & 34.65 & 25.39 & 21.93 \\
MonoCon + \textbf{Ours}      & 23.47 & 17.07 & 14.20 & 27.13 & 20.57 & 17.40 & 36.31 & 26.60 & 22.87 \\
Improvement                  & \textbf{+0.97} & \textbf{+0.61} & \textbf{+0.25} & \textbf{+0.8} & \textbf{+1.56} & \textbf{+1.42} & \textbf{+1.66} & \textbf{+1.21} & \textbf{+0.94} \\
\hline 
MonoDETR \cite{zhang2022monodetr}                    & 23.65 & 15.92 & 12.99 & 28.84 & 20.61 & 16.38 & 37.86 & 26.95 &  22.80\\
MonoDETR + \textbf{Ours}     & 24.54 & 17.14& 14.59& 32.24& 22.06 & 18.50 & 41.82 & 29.33 & 25.23   \\
Improvement                  &\textbf{+0.89}& \textbf{+1.22} & \textbf{+1.60} &  \textbf{+3.40} & \textbf{+1.45} & \textbf{+2.12} & \textbf{+3.96} & \textbf{+2.38} & \textbf{+2.43}  \\
\hline
\end{tabular}
}

\label{comapre with base models}
\end{table*}

\begin{table*}[t]
\centering
\caption{\textbf{Comparison to recent SOTA models.} The best results are in \textbf{bold}. Our approach consistently achieves the best accuracy on both the KITTI validation and test splits.}
\setlength{\tabcolsep}{3.5mm}
{
\begin{tabular}{c|c|ccc|ccc|ccc}
\hline
\multirow{2}{*}{ Method } & \multirow{2}{*}{ Reference } &\multicolumn{3}{c|}{$A P_{3 D}$ (Car test) } & \multicolumn{3}{c|}{$A P_{B E V}$ (Car test) } & \multicolumn{3}{c}{$A P_{3 D}$ (Car val) } \\
\cline{3-11}
& & Mod. & Easy & Hard & Mod. & Easy & Hard & Mod. & Easy & Hard \\
\hline 
CMAN \cite{cao2022cman}  & TITS2022 & 11.87 & 17.77 & 9.16 & 17.04 & 25.89 & 12.88 & 16.75 & 24.91 & 12.46 \\
MonoDLE \cite{ma2021delving}  & CVPR2021 & 12.26 & 17.23 & 10.29 & 18.89 & 24.79 & 16.00 & 13.66 & 17.45 & 11.68 \\
MonoRUn \cite{chen2021monorun} & CVPR2021 & 12.30 & 19.65 & 10.58 & 17.34 & 27.94 & 15.24 & 14.65 & 20.02 & 12.61 \\
MonoRCNN \cite{shi2021geometry}  & ICCV2021 & 12.65 & 18.36 & 10.03 & 18.11 & 25.48 & 14.10 & 14.87 & 19.07 & 12.59 \\
DDMP-3D \cite{wang2021depth} & CVPR2021 & 12.78 & 19.71 & 9.80 & 17.89 & 28.08 & 13.44 & - & - & - \\
HMF \cite{liu2022fine}  &  TIP2022 & 13.12 & 20.28 & 9.56 & - & - & - & - & - & - \\

MonoPixel \cite{kim2022boosting} & TITS2023 & 13.33 & 19.06 & 11.90 & 19.75 & 27.98 & 17.32 & - & - & - \\
CaDDN \cite{reading2021categorical} & CVPR2021 & 13.41 & 19.17 & 11.46 & 18.91 & 27.94 & 17.19 & 16.31 & 23.57 & 13.84 \\
MonoEF \cite{zhou2021monoef} & TPAMI2021 & 13.87 & 21.29 & 11.71 & 19.70 & 29.03 & 17.26 & 16.30 & 18.26 & 15.24 \\
MonoFlex \cite{zhang2021objects}  & CVPR2021 & 13.89 & 19.94 & 12.07 & 19.75 & 28.23 & 16.89 & 17.51 & 23.64 & 14.83 \\
AutoShape \cite{liu2021autoshape} & ICCV2021 & 14.17 & 22.47 & 11.36 & 20.08 & 30.66 & 15.59 & 14.65 & 20.09 & 12.07 \\
MonoGround \cite{qin2022monoground} & CVPR2022 & 14.36 & 21.37 & 12.62 & 20.47 & 30.07 & 17.74 & 18.69 & 25.24 & 15.58  \\
DEVIANT \cite{kumar2022deviant} & ECCV2022 & 14.46 & 21.88 & 11.89 & 20.44 & 29.65 & 17.43 & 16.54 & 24.63 & 14.52 \\

Context-Aware \cite{zhou2022context}  & TITS2022 & 14.49 & 20.89 & 12.19 & 20.77 & 29.57 & 17.88 & 16.09 & 20.15 & 15.59 \\

Homo \cite{gu2022homography} & CVPR2022 & 14.94 & 21.75 & 13.07 & 20.68 & 29.60 & 17.81 & 16.89 & 23.04 & 14.90 \\
GupNet \cite{lu2021geometry} & ICCV2021 & 15.02 & 22.26 & 13.12 & 21.19 & 30.29 & 18.20 & 16.46 & 22.76 & 13.72 \\
MonoPoly \cite{guan2022monopoly}  & PR2022 & 15.22 & 22.07 & 13.24 & 21.08 & 30.32 & 18.19 & 16.13 & 22.94 & 13.01 \\
MonoDTR \cite{huang2022monodtr} & CVPR2022 & 15.39 & 21.99 & 12.73 & 20.38 & 28.59 & 17.14 & 18.57 & 24.52 & 15.51 \\
DCD \cite{li2022densely} & ECCV2022 & 15.90 & 23.81 & 13.21 & 21.50 & 32.55 & 18.25 & 17.38 & 23.94 & 15.32 \\
MonoDETR \cite{zhang2022monodetr} & ICCV2023 & 15.92 & 23.65 & 12.99 & 21.44 & 32.08 & 17.85 & 20.61 & 28.84 & 16.38 \\
Monodistill \cite{chong2022monodistill} & ICLR2022 & 16.03 & 22.97 & 13.60 & 22.59 & 31.87 & 19.72 & 18.47 & 24.31 & 15.76 \\
PDR  \cite{sheng2023pdr}  &  TCSVT2023 & 16.14 & 23.69 & 13.78 & 21.74 & 31.76 & 18.79 & 19.44 & 27.65 & 16.24 \\
Monocon \cite{liu2022learning}  & AAAI2022 & 16.46 & 22.50 & 13.95 & 22.10 & 31.12 & 19.00 & 19.01 & 26.33 & 15.98 \\
Shape-Aware \cite{chen2023shape}  & TITS2023 & 16.52 & 23.84 & 13.88 & - & - & - & 18.39 & 24.92 & 15.56 \\ 
\hline 
Monocon+\textbf{Ours} & - & 17.07 & 23.47 & 14.20 & \textbf{24.47} & \textbf{33.35} & \textbf{20.12} & 20.57 & 27.13 & 17.40 \\
MonoDETR+\textbf{Ours}  & - & \textbf{17.14} & \textbf{24.54} & \textbf{14.59} & 22.42 & 32.19 & 19.48 & \textbf{22.06} & \textbf{32.24} & \textbf{18.50} \\
\hline
\end{tabular}
}

\label{comapre with sotas}
\end{table*}

\textbf{3) Training Settings. }
For the KITTI dataset, our approach is trained end-to-end for 200 epochs and adopts a warm-up strategy of 5 epochs. For the NuScenes dataset, the total epoch is 24. For MonoDLE*, we employ Adam optimizer with an initial learning rate of $1.25e^{-3}$, and decay it by ten times at 120 and 160 epochs. Data augmentation consists of random flips and center crops. For MonoCon, the common Adam optimizer is utilized with an initial learning rate of $2.25e^{-4}$, with a cyclic scheduler. Data augmentation consists of random flips and center crops. For MonoDETR, the AdamW optimizer is used with an initial learning rate of $2e^{-4}$, and decays it by ten times at 125 and 165 epochs. Data augmentation consists of random flips, center crops, and random shifts. For FCOS3D*, the SGD optimizer is used with an initial learning rate of $2e^{-3}$, and decays it by ten times at 16 and 22 epochs. Data augmentation consists of random flips.

\subsection{Evaluation}

In this section, we compare the performance of our approach with dozens of recent baselines on both the KITTI validation and test sets. Firstly, we compare our approach with the three base models in Table \ref{comapre with base models}. Secondly, we compare our approach with other recent SOTA methods in Table \ref{comapre with sotas}. Then, we compare our approach with other cross-modality distillation methods in Table \ref{comapre with monodistill}. Finally, we additionally present the results of our approach on pedestrian and bicycle categories in Table \ref{Pedestrian and Cyclist}.

\textbf{1) Comparison to Three Base Models On KITTI.}
Table \ref{comapre with base models} shows that our approach consistently outperforms the three base models across all difficulty levels on both the KITTI validation and test sets. Specifically, our approach achieves significant gains on MonoDLE* and also well improves the accuracy of MonoDETR and Monocon on the validation split. The gains on MonoDETR and Monocon are lower than those on MonoDLE*. One reason lies that MonoCon incorporated various auxiliary tasks, which partly alleviates the impact of modality gap. In addition, the architecture of MonoDETR is not suitable for LiDAR inputs. Notably, these two models achieve SOTA accuracy among all released models of monocular 3D detection, so improving their accuracy is more challenging. Despite these challenges, our approach still achieves a considerable improvement on all these base models.

\begin{table}[t]
\centering
\caption{\textbf{Comparison to Monodistill with the same base model on KITTI.} The results show that our approach more effectively alleviates feature overfitting issue in cross-modality distillation. The gain of our approach is mainly attributed to the developed DASFD and DASRD modules.}
\setlength{\tabcolsep}{1.0mm}{
\begin{tabular}{c|ccc|ccc}
\hline 
\multirow{2}{*}{ Method } & \multicolumn{3}{c}{$AP_{3D}$ (Car test) } & \multicolumn{3}{|c}{$AP_{3D}$( Car val) } \\
\cline{2-7}
& Easy & Mod. & Hard & Easy & Mod. & Hard  \\ 
\hline
MonoDLE* & 17.23 & 12.26 & 10.29 & 19.29 & 15.13 & 12.78 \\
MonoDistill & 22.97 & 16.03 & 13.60 & 24.31 & 18.47 & 15.76  \\
MonoDLE*+\textbf{Ours} & 24.04 & 16.50 & 13.85 & 25.59 & 19.23 & 16.60  \\
\hline 
vs Monodistill & \textbf{+1.07} & \textbf{+0.47} & \textbf{+0.25} & \textbf{+1.28 }& \textbf{+0.76 }& \textbf{+0.84}  \\
\hline
\end{tabular}
}

\label{comapre with monodistill}
\end{table}

\textbf{2) Comparison to Recent SOTA Models.}
Table \ref{comapre with sotas} shows the comparison of our approach and dozens of released SOTA models in recent years. Benefiting from the effectiveness of the developed \textbf{DASFD} and \textbf{DASRD} modules as well as the strong ability of the base models, our approach expectedly achieves the best accuracy compared with all these baseline models. Notably, our approach serves as a general distillation framework that can be applied to various CNN-based and DETR-based detection models. Therefore, it can be seamlessly integrated into stronger base models to achieve higher accuracy without increasing inference cost once their codes are released in the future.

\textbf{3) Comparison using the Same Base Model.}
We further compare our approach with existing cross-modality distillation methods. we select the recent Monodistill for comparison. Both our approach and Monodistill employ MonoDLE* as the base model for a fair comparison. In this experiment, we employ the same base model as Monodistill and avoid the impact of the architecture inconsistency issue. Therefore, the accuracy gap between the two approaches primarily arises from the feature overfitting issue. Table \ref{comapre with monodistill} shows that our approach considerably outperforms Monodistill on both the training and validation sets. It indicates that our approach effectively alleviates the feature overfitting issue in cross-modality distillation while Monodistill does not well handle this problem. The improvement of our approach is mainly attributed to the developed \textbf{DASFD} and \textbf{DASRD} modules, which alleviate the feature overfitting issue by selectively learning positive features and relationships of objects in cross-modality distillation.

\textbf{4) Comparison for Pedestrian and Cyclist.}
Detecting pedestrians and cyclists is more challenging compared to cars due to their small sizes and limited training samples. Similar to other methods, our approach focuses on detecting cars while maintains competitive performance on pedestrians and cyclists categories. Therefore, the knowledge distillation is mainly conducted for the Car category in our approach. Table \ref{Pedestrian and Cyclist} presents the detection results of pedestrians and cyclists categories on the KITTI test set. It is seen that our approach based on MonoDLE* still achieves the best accuracy across all difficulty levels on the pedestrians category and competitive accuracy on the cyclist category compared with recent models. The detection gain on these categories partly benefits from the moderate assistance of LiDAR data in our approach. Notably, the training set in our approach contains few samples for pedestrians and cyclists categories. It remains a challenging task in our future work.

\begin{table}
\centering
\caption{\textbf{Comparison for pedestrian and cyclist categories on the KITTI test set with IOU thresholds of 0.5.} Our approach achieves the best accuracy across all difficulty levels on pedestrian and competitive accuracy on cyclist.}
\setlength{\tabcolsep}{2.2mm}{
\begin{tabular}{c|ccc|ccc}
        \hline
        \multirow{2}{*}{Method} & \multicolumn{3}{c|}{$AP_{3D}$ (Pedestrian)} & \multicolumn{3}{c}{$AP_{3D}$ (Cyclist)} \\
        \cline{2-7}
        & Easy & Mod. & Hard & Easy & Mod. & Hard \\
        \hline
        MonoEF & 4.27 & 2.79 & 2.21 & 1.80 & 0.92 & 0.71 \\
        D4LCN & 4.55 & 3.42 & 2.83 & 2.45 & 1.67 & 1.36 \\
        M3D-RPN & 4.92 & 3.48 & 2.94 & 0.94 & 0.65 & 0.47 \\
        DDMP-3D & 4.93 & 3.55 & 3.01 & 4.18 & 2.50 & 2.32 \\
        AutoShape & 5.46 & 3.74 & 3.03 & 5.99 & 3.06 & 2.70 \\
        MonoGeo & 8.00 & 5.63 & 4.71 & 4.73 & 2.93 & 2.58 \\
        MonoFlex & 9.43 & 6.31 & 5.26 & 4.17 & 2.35 & 2.04 \\
        MonoDLE & 9.64 & 6.55 & 5.44 & 4.59 & 2.66 & 2.45 \\
        MonoPair & 10.02 & 6.68 & 5.53 & 3.79 & 2.21 & 1.83 \\
        \hline
        \textbf{MonoDLE*+Ours} & 10.23 & 6.91 & 5.80 & 4.15&2.47 &2.17 \\
        \hline
\end{tabular}
}
\label{Pedestrian and Cyclist}
\end{table}

\subsection{Ablation Studies}

In this section, we conduct ablation studies to confirm the effectiveness of different modules and influence of parameters in our approach. First, we verify the effectiveness of our distillation modules \textbf{DASFD} and \textbf{DASRD} on both the KITTI and NuScenes datasets in Table \ref{ablation} and table \ref{ablation on NuScenes}. 
Second, we verify the criterion of depth uncertainty in our approach with a general criterion in Table \ref{Comparisons different criterions for selective learning}.
Third, we further compare different weight schemes of depth uncertainty estimation in Table \ref{Comparisons different weight scheme}.  
Last, we analyze the effect of hyperparameters and the teacher network in Tables \ref{tab:hyperparameters} and \ref{Comparisons Different teacher models}, respectively.

\begin{table}[t!]
\centering
\caption{ \textbf{Ablation studies of distillation modules on KITTI validation.} FD, RD, and ED refer to general feature, relation, and response distillation. DASFD and DASRD refer to our distillation and relation distillation.}
\setlength{\tabcolsep}{1.5mm}{
\begin{tabular}{c|cc|ccc|ccc}
\hline
\multirow{2}{*}{\#} & \multicolumn{2}{c|}{ General } & \multicolumn{3}{c|}{ Ours }  & \multicolumn{3}{c}{$AP_{3D}$ (Car val) } \\
\cline{2-9} 
& FD & RD & DASFD  & DASRD & ED & Easy & Mod. & Hard \\
\hline 
a. & & & &  & & 19.29 & 15.13 & 12.78 \\
\hline
b. & & & &  & $\checkmark$ & 21.63 & 17.24 & 14.71 \\
\hline 
c. & $\checkmark$ & $\checkmark$ &  & & $\checkmark$ & 24.31 & 18.47 & 15.76 \\
d. & $\checkmark$ & &  & $\checkmark$ & $\checkmark$ & 24.78 & 18.57 & 16.50 \\
e. & & $\checkmark$ & $\checkmark$ &  & $\checkmark$ & 25.37 & 19.06 & 16.33 \\
f. & & & $\checkmark$ &  $\checkmark$ & $\checkmark$ & \textbf{25.59} & \textbf{19.23} & \textbf{16.60} \\
\hline
\end{tabular}
}

\label{ablation}
\end{table}

\begin{table}[t!]
\centering
\caption{ \textbf{Ablation studies of distillation modules on NuScenes validation.} FCOS3D* refers to the results of its open-source model. Notably, all methods don’t utilize test-time augmentation (TTA) and model ensemble.}
\setlength{\tabcolsep}{2.0mm}{
\begin{tabular}{c|c|c|cc}
\hline
Method & DASFD & DASRD & MAP↑ & NDS↑ \\
\hline 
FCOS3D* & & & 0.321& 0.395 \\
FCOS3D*+DASFD & \checkmark &  &0.331 &0.400  \\
FCOS3D*+DASRD & & \checkmark &0.328 & 0.396 \\
FCOS3D*+\textbf{Ours} & \checkmark & \checkmark &\textbf{ 0.335}&\textbf{ 0.400} \\
\hline
\end{tabular}
}

\label{ablation on NuScenes}
\end{table}

\textbf{1) Effectiveness of Distillation Modules on KITTI.}
Our approach mainly comprises two novel distillation modules including \textbf{DASFD} and \textbf{DASRD}. Table \ref{ablation} shows the ablation studies of each individual module, as well as the general response distillation. MonoDLE* is adopted as the base model. First, Row a and b simply verify the effectiveness of the individual response distillation. Second, Row d and c verify the effectiveness of the individual \textbf{DASRD}. Third, Row e and c verify the effectiveness of the individual \textbf{DASFD}. Last, our final solution achieves a significant performance improvement, when all three distillation modules are employed in Row f. In conclusion, both the two modules \textbf{DASFD} and \textbf{DASRD} in our approach work effectively to alleviate the negative transfer problem in cross-modality distillation.

\textbf{2) Effectiveness of Distillation Modules on NuScenes.} We further verify the effectiveness of our two distillation modules on another dataset NuScenes. To this end, we integrate our approach into the base model FCOS3D* \cite{wang2021fcos3d}, which was well implemented on NuScenes by the authors. Table \ref{ablation on NuScenes} demonstrates that both the two individual modules \textbf{DASFD} and \textbf{DASRD } could well improve the accuracy of the base model. Our final solution achieves a significant performance improvement, when both the two distillation modules are employed.

\begin{table}[t!]
\centering
\caption{\textbf{Ablation study on depth uncertainty. } Employing depth uncertainty consistently achieves better results across all difficult levels compared with depth error.}
\setlength{\tabcolsep}{5.0mm}{
\begin{tabular}{c|ccc}
        \hline
        \multirow{2}{*}{Method} & \multicolumn{3}{c}{$AP_{3D}$ (Car val)}  \\
        \cline{2-4}
        & Easy & Mod. & Hard  \\
        \hline
        w/ depth error & 24.27 & 18.36 & 15.59  \\
        w/ depth uncertainty & 25.59 & 19.23 & 16.60 \\
        \hline
\end{tabular}
}

\label{Comparisons different criterions for selective learning}
\end{table}

\begin{table}[t!]
\centering
\caption{\textbf{Ablation studies on weight schemes of depth uncertainty.} Both the two fusion strategies yield even worse results than a separate weight scheme.}
\setlength{\tabcolsep}{3.5mm}{
\begin{tabular}{c|c|ccc}
        \hline
        \multirow{2}{*}{} & \multirow{2}{*}{Scheme} & \multicolumn{3}{c}{$AP_{3D}$ (Car val)}  \\
        \cline{3-5}
        & & Easy & Mod. & Hard  \\
\hline 
\multirow{2}{*}{separately} & Student & 25.59 & 19.23 & 16.60 \\
& Teacher & 24.65 & 19.14 & 16.45 \\
\hline 
\multirow{2}{*}{fusion} & sum & 24.04 & 18.12 & 15.47 \\
& multiply & 24.37 & 18.42 & 16.41 \\
        \hline
\end{tabular}
}

\label{Comparisons different weight scheme}
\end{table}

\textbf{3) Ablation Study on Depth Uncertainty.}
Our approach requires an effective evaluation metric for selective learning in cross-modality distillation. Both depth errors and depth uncertainty could well reflect the depth prediction capability of the student network. Table \ref{Comparisons different criterions for selective learning} compares the performance of our approach using depth uncertainty and depth errors as the criterion. Employing depth uncertainty consistently achieves better results across all difficulty levels compared with depth errors. The reason lies that distant objects often exhibit larger depth errors compared with nearby objects. It results in the imbalance issue of objects when using depth errors as the evaluation metric. In contrast, depth uncertainty well reflects the confidence of the network in its predictive capability.

\textbf{4)  Weighting Schemes of depth uncertainty.} In our solution, depth uncertainty is obtained from the student network, denoted "student". It can be optionally obtained from the teacher network, or a fusion of the teacher and student networks. 
When depth uncertainty is obtained from the teacher network, denoted "teacher", it is calculated as $\omega_i=1-\sigma_{T, i}$. In this way, the teacher network assigns higher weight values to well-learned objects and lower weight values to poorly-learned objects. It ensures that depth information is only provided by high-quality objects to avoid the negative transfer problem for the student network, and vice versa. Depth uncertainty can also be obtained from a fusion of the teacher and student networks. We take two operations for fusion including summation and multiplication of these two schemes, namely "sum" and "multiply". 
Overall, the "student" scheme performs better than others in Table \ref{Comparisons different weight scheme}. The "teacher" scheme tends to prioritize higher-quality objects, while the "student" scheme often focuses on objects that are not well predicted. The latter is more suitable for distillation because it can transfer more positive information for low-quality objects. As a result, the fusion of the two weight schemes yields even worse accuracy than a separate weight scheme. 

\begin{table}[t!]
    \centering
    \caption{\textbf{Ablation Studies of the loss weight $\lambda_3$}. The loss weights $\lambda_1$ and $\lambda_2$ have been set to be 10 and 1 following Monodistil \cite{chong2022monodistill}.}
    \setlength{\tabcolsep}{4.0mm}
    \renewcommand{\arraystretch}{1.2}
    \begin{tabular}{c|c|ccc}
        \hline
        \multirow{2}{*}{Method} & \multirow{2}{*}{$\lambda_3$} & \multicolumn{3}{c}{$AP_{3D}$ (Car val)} \\
        \cline{3-5}
         & & Easy & Mod. & Hard \\
        \hline
        \multirow{3}{*}{MonoDLE*} & 0.1 & 21.68 & 17.04 & 14.55 \\
         & 1 & \textbf{25.59} & \textbf{19.23} & \textbf{16.60} \\
        & 10 & 22.48 & 16.65 & 14.31 \\
        \hline
        \multirow{3}{*}{MonoDETR} & 0.1 & \textbf{32.24} & \textbf{22.06} & \textbf{18.50} \\
         & 1 & 29.59 & 21.58 & 18.06 \\
        & 10 & 30.71 & 21.25 & 17.73 \\
        \hline
    \end{tabular}
    
    \label{tab:hyperparameters}
\end{table}

\begin{table}[t!]
\centering
\caption{\textbf{Effect of the teacher network.} Our approach is not highly influenced by the accuracy of the teacher network.}
\setlength{\tabcolsep}{3.5mm}{
\begin{tabular}{c|c|ccc}
        \hline
        \multirow{2}{*}{Method} & \multirow{2}{*}{network} & \multicolumn{3}{c}{$AP_{3D}$ (Car val)} \\
        \cline{3-5}
        & & Easy & Mod. & Hard  \\
        \hline

\multirow{3}{*}{Monocon}& Teacher & 68.13 & 49.72 & 41.48  \\

& Base model  & 26.33 & 19.01 & 15.98  \\
& Ours & 27.13 & 20.57 & 17.40  \\
\hline
\multirow{3}{*}{MonoDETR} & Teacher  & 40.07 & 27.68 & 22.49  \\

  & Base model & 28.84 & 20.61 & 16.38   \\
& Ours & 32.24 & 22.06 & 18.50   \\
        \hline
\end{tabular}
}

\label{Comparisons Different teacher models}
\end{table}

\begin{figure*}
    \centering
    \includegraphics[width=17cm]{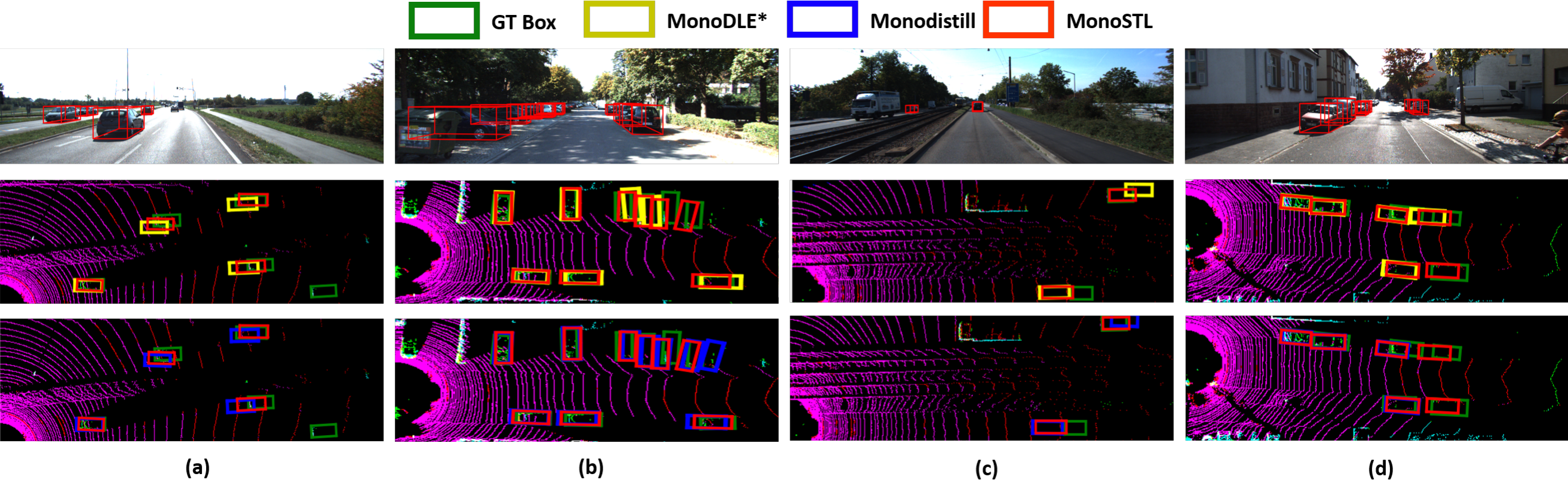}
    \caption{\textbf{Visual results in BEV view}. The first row in each example displays 3D boxes detected by our approach on images. The second row compares our approach with the base model MonoDLE* in BEV view. The third row compares our approach with the recent Monodistill using the same base model in BEV view. The improvement of our approach is mainly attributed to the developed \textbf{DASFD} and \textbf{DASRD} modules.}
    \label{fig:visual_compare}
\end{figure*}

\begin{figure*}
    \centering
    \includegraphics[width=18cm]{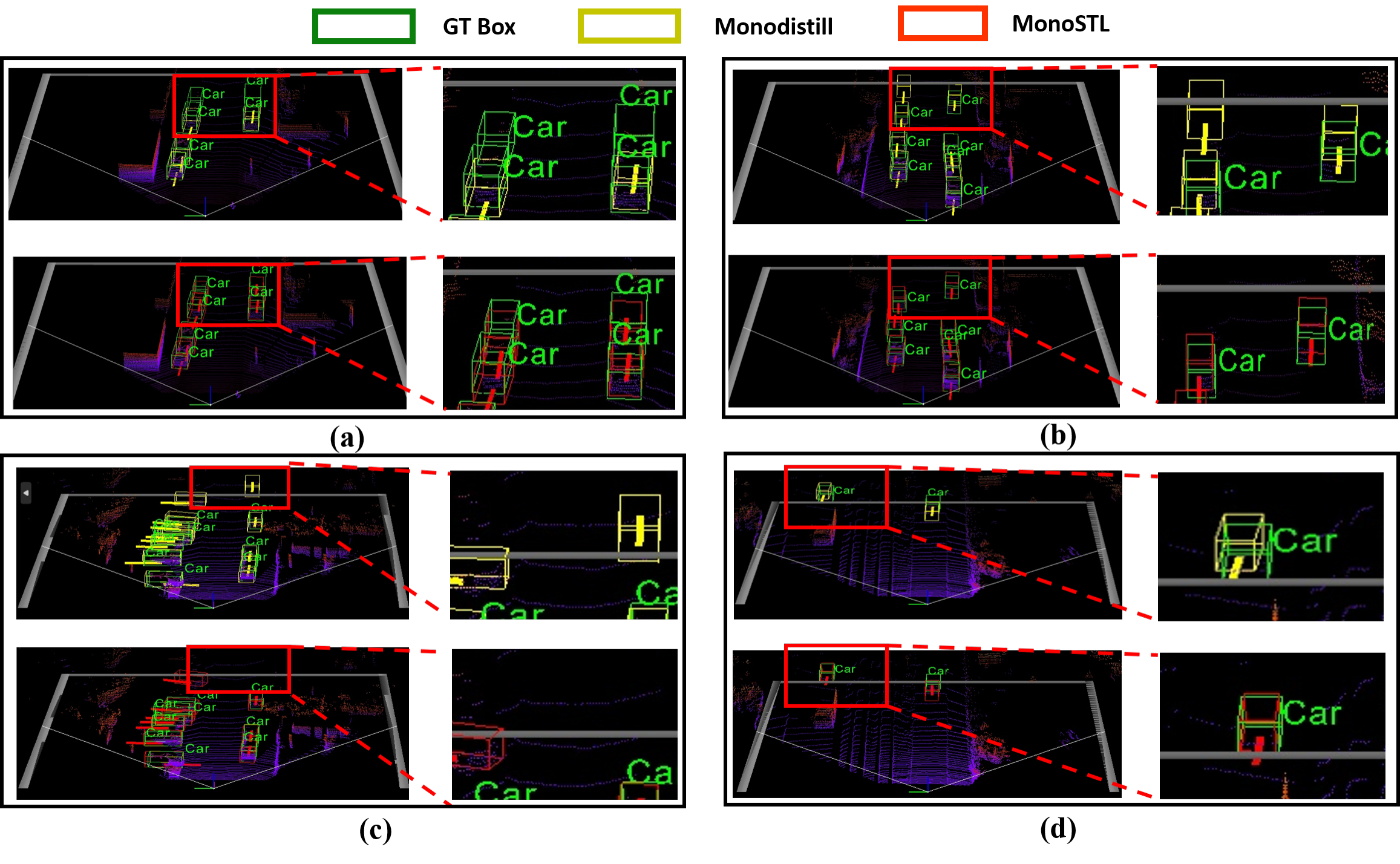}
    \caption{ \textbf{Visual results in lateral view}. The first and second col in each example displays 3D boxes detected by our approach and Monodistill, respectively. Our approach performs better than Monodistill in cases of false positives and false negatives.}
    \label{fig:front_3d_box_visual}
\end{figure*}

\textbf{5) Ablation Study on Loss Weights.}
The hyperparameters $\lambda_1$, $\lambda_2$, and $\lambda_3$ in equation (\ref{total loss}) are utilized to balance different losses in our approach. The two loss weights for feature distillation ($\lambda_1$) and relation distillation ($\lambda_2$) have been set to be 10 and 1 following  the settings in Monodistill. Table \ref{tab:hyperparameters} shows the influence of the loss weight for response distillation $\lambda_3$. The CNN-based MonoDLE* and the DETR-based MonoDETR are adopted as base models. Based on the ablation study, the value of the loss weight $\lambda_3$ is set to 1 for CNN-based models and 0.1 for DETR-based models in our approach. Notably, Table \ref{Comparisons Different teacher models} shows that the accuracy of the teacher model for MonoDETR is much lower than MonoDLE* and MonoCon. It further indicates that a smaller loss weight $\lambda_3$ for response distillation is more suitable for DETR-based models.

\textbf{6) Impact of the Teacher Network.}
Our approach leverages effective LiDAR-based detectors for monocular 3D detection. Is our approach highly influenced by the accuracy of LiDAR-based detectors? Table \ref{Comparisons Different teacher models} shows the accuracy of the teacher networks for the base models MonoCon and MonoDETR in our approach. Notably, these two recent models achieve comparable accuracy of monocular 3D detection. It is seen that the accuracy of the teacher network for MonoDETR is much lower than MonoCon across all difficulty levels. The reason lies that MonoCon employs LiDAR as input, while MonoDETR uses both LiDAR and images. Therefore, input LiDAR only affects the depth encoder of MonoDETR, which limits its accuracy compared with MonoCon. Nevertheless, our approach achieves desired results on both MonoCon and MonoDETR. Indeed, our approach on MonoDETR achieves the best accuracy on both validation and test sets using the metric AP3D than all released SOTA models. It indicates that our approach is not highly influenced by the accuracy of the teacher network. The effectiveness of cross-modality distillation is mainly determined by whether the model successfully transfers positive information while blocking interference from negative information.

\begin{figure}[t!]
    \centering
    \includegraphics[width=8cm]{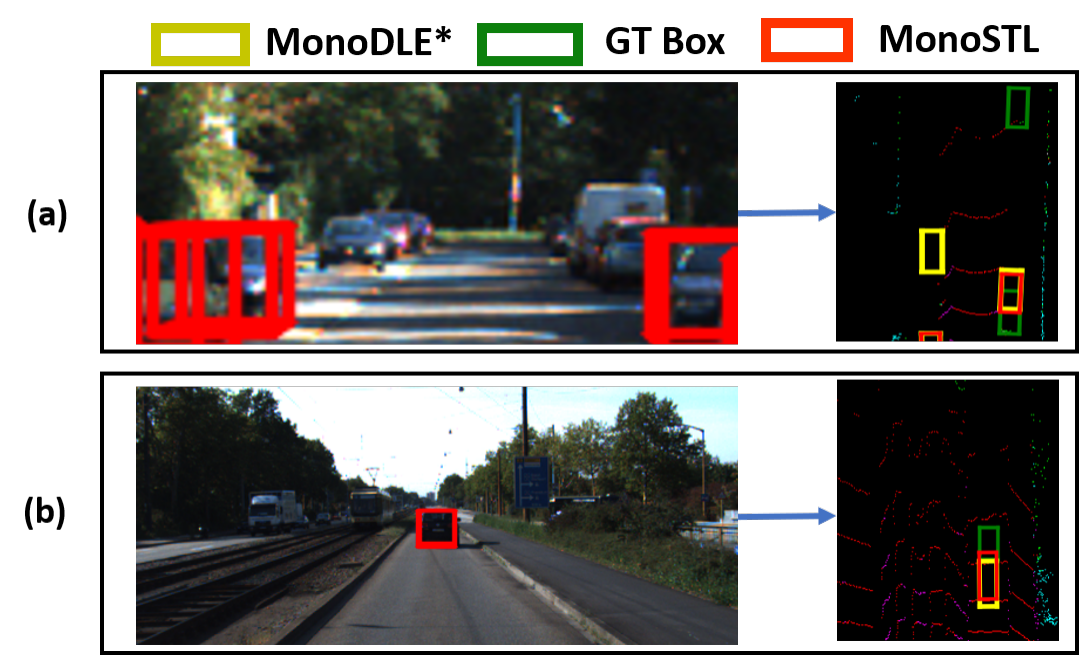}
    \caption{\textbf{Failure examples.}  A few objects are missed or detected inaccurately by our approach. The failure case is mainly caused by the inaccurate depth estimation of object centers.
    }
    \label{fig:bad_cases}
\end{figure}

\begin{figure}[t!]
    \centering
    \includegraphics[width=9cm]{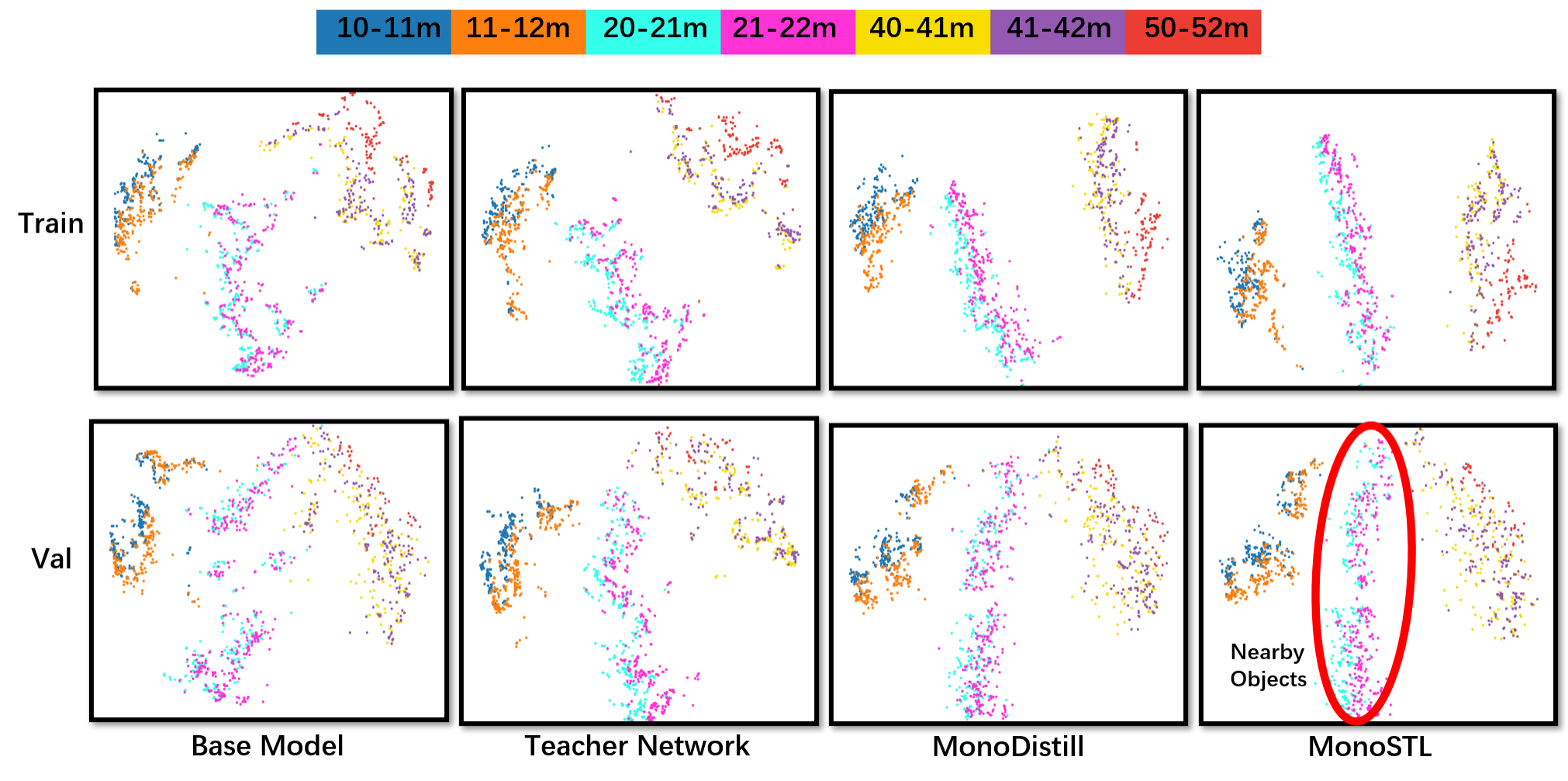}
    \caption{\textbf{t-SNE visualization of network features on both train and validation sets.} Nearby objects, i.e. objects with close GT labels, are distinguished more easily in our approach than the base model and monodistill, such as blue (20-21m) and pink (21-22m). It indicates that the feature overfitting issue in our approach is well alleviated than others.
    }
    \label{fig:t_SNE}
\end{figure}

\subsection{Visualization}

In this section, we show visual results of detected 3D boxes to clearly demonstrate the effectiveness of our approach. We also offer failure examples and analysis. In addition, we adopt t-SNE feature analysis to intuitively show the effectiveness of our approach in alleviating the negative transfer problem.

\textbf{1) Visual Results in BEV view.} 
Fig. \ref{fig:visual_compare} shows the visual comparison of our approach with the base model MonoDLE* and the recent cross-modality distillation method Monodistill. Both our approach and Monodistill employ
MonoDLE* as the base model for a fair comparison. It is clear that our approach achieves superior detection performance compared to the base model and Monodistill, especially for distant and occluded objects. Though Monodistill improves the detection results compared to the base model, it also introduces false positives while reducing false negatives in Fig. \ref{fig:visual_compare}(b). It confirms that Monodistill indiscriminately learns depth information from the teacher network, which helps in detecting distant targets but also results in false positives due to feature overfitting. In contrast, our approach selectively learns positive features from the teacher network, resulting in better detection results and reducing false positives and false negatives.

\textbf{2) Visual Results in lateral view.} 
Fig. \ref{fig:front_3d_box_visual} further shows the visual results of our approach compared with the recent cross-modality distillation approach Monodistill in lateral view. In Fig. \ref{fig:front_3d_box_visual}(a) and Fig. \ref{fig:front_3d_box_visual}(b), our approach successfully detected objects missed by Monodistill. In Fig. \ref{fig:front_3d_box_visual}(c), false detections observed in Monodistill are avoided in our approach. In addition, in Fig. \ref{fig:front_3d_box_visual}(d), our approach localizes distant objects more accurately than Monodistill. The improvement of our approach is attributed to the developed \textbf{DASFD} and \textbf{DASRD} modules, which well alleviate the feature overfitting issue in cross-modality distillation. 

\textbf{3) Failure examples.} 
We observe in Fig. \ref{fig:bad_cases} that a few objects are occasionally missed or detected inaccurately. In fig. \ref{fig:bad_cases}(a), both the base model and our approach miss the farthest object. In fig. \ref{fig:bad_cases}(b), our approach does not locate the object accurately. It indicates that, despite providing depth information to images through cross-modality distillation, the network still faces challenges in overcoming the ill-posed problem of monocular depth estimation.

\textbf{4) t-SNE Feature Visualization and Analysis.}
T-distributed stochastic neighbor embedding (t-SNE) \cite{hinton2002stochastic,van2008visualizing} is a statistical method for visualizing high-dimensional data. We visualize network features using t-SNE on the head features used for depth prediction. The base model MonoDLE* and the recent Monodistill are used for comparison. We use t-SNE to reduce the dimension of depth prediction features to two dimensions and use those as x and y coordinates to visualize the distributions of depth features. In Fig. \ref{fig:t_SNE}, each point corresponds to an object. Different colors represent the ranges of objects using GT depth labels. Objects with closer GT depth labels are referred as nearby objects.

When two objects in the figure are plotted more closely, it may be more challenging to distinguish their depth. Consequently, when nearby objects in the figure exhibits greater separation, it indicates stronger depth prediction performance of the model. Additionally, if a model effectively distinguishes nearby objects on both the training and validation sets, it indicates effective generalization ability of the model. Conversely, if a model performs well on the training set but not well on the validation set, it may suffer from the feature overfitting issue. The detailed analysis is as follows:

\begin{itemize}
    \item \textbf{Feature distributions w/ and w/o distillation.} 
    Before distillation, feature distributions of the base model are clearly inconsistent on train and validation sets. It indicates that learned features from training set are ineffective on validation sets for monocular 3D detection due to the lack of depth information in images. After distillation by our approach, feature distributions of the student network are well consistent with each other on train and validation sets than base model. Additionally, nearby objects could be easily distinguished by our approach than the base model. It indicates that cross-modality distillation effectively transfers missing depth information from LiDAR data to the student network, resulting in the performance improvement of monocular 3D detection.
    
    \item \textbf{How our approach alleviates feature overfitting?} 
    Existing cross-modality distillation methods such as Monodistill could easily distinguish nearby objects in the train set. However, they do not work well in the validation set due to the feature overfitting issue caused by modality gap.
    By comparison, our approach could well distinguish nearby objects in both train and validation sets. It ensures the generalization ability of monocular 3D detection on validation and test sets. The improvement of our approach is mainly beneficial to the selective learning strategy from the teacher network to alleviate the feature overfitting issue in cross-modality distillation.

\end{itemize}

\section{Conclusion}
This paper systematically investigated the negative transfer problem of cross-modality distillation in monocular 3D detection including not only the architecture inconsistency issue but more importantly the feature overfitting issue. The proposed method was seamlessly integrated into four base models to considerably improve the detection accuracy on multiple datasets. It also held the potential to achieve higher accuracy with stronger CNN-based and DETR-based models in the future. In addition, the proposed method provides a fundamental solution for cross-modality distillation between multi-modal RGB and depth/LiDAR for other relevant tasks in this field.

Though our approach has well alleviated the negative transfer problem for monocular 3D detection, how to completely eliminate the modality gap in cross-modality distillation still remains a challenge. On the one hand, existing methods to resolve the architecture inconsistency issue are not always applicable to all base models. On the other hand, it is still unable to exactly determine which transferred features are beneficial in cross-modality distillation.

\bibliographystyle{IEEEtran}

\end{document}